\documentclass[runningheads]{llncs}
\usepackage{graphicx}
\usepackage{microtype}
\usepackage{multirow}
\usepackage{amssymb}
\usepackage{pifont}
\newcommand{\cmark}{\ding{51}}%
\newcommand{\xmark}{\ding{55}}%
\usepackage{adjustbox}
\usepackage{array, makecell}

\begin{document}
\title{PDF-VQA: A New Dataset for Real-World VQA on PDF Documents}

\author{Yihao Ding\inst{1}\thanks 
{
 ~~Co-First Authors
} \and
Siwen Luo\inst{1*},
Hyunsuk Chung\inst{3}, Soyeon Caren Han  \inst{1,2}\thanks{
~~Corresponding Author}}
\authorrunning{Y. Ding et al.}
%
\institute{Unversity of Sydney, Sydney NSW 2006 \email{\{yihao.ding,siwen.luo,caren.han\}@sydney.edu.au}\and
The University of Western Australia, Perth WA 6009
\email{caren.han@uwa.edu.au} \and FortifyEdge, Sydney, Australia, NSW \email{davidchungproject@gmail.com}}




\maketitle              
\begin{abstract}
Document-based Visual Question Answering examines the document understanding of document images in conditions of natural language questions. We proposed a new document-based VQA dataset, PDF-VQA, to comprehensively examine the document understanding from various aspects, including document element recognition, document layout structural understanding as well as contextual understanding and key information extraction. Our PDF-VQA dataset extends the current scale of document understanding that limits on the single document page to the new scale that asks questions over the full document of multiple pages. We also propose a new graph-based VQA model that explicitly integrates the spatial and hierarchically structural relationships between different document elements to boost the document structural understanding. The performances are compared with several baselines over different question types and tasks\footnote{The full dataset will be released after paper acceptance. The partial dataset is provided in the supplementary material.}.

\keywords{Document Understanding \and Document Information Extraction \and Visual Question Answering}
\end{abstract}
\section{Introduction}

With the rise of digital documents, document understanding received much attention from leading industrial companies, such as IBM \cite{zhong2019publaynet} and Microsoft \cite{xu2020layoutlm,xu2021layoutlmv2}. Visual Question Answering (VQA) on visually-rich documents (i.e. scanned document images or PDF file pages) aims to examine the comprehensive document understandings in conditions of the given questions \cite{kafle2018dvqa}. A comprehensive understanding of a document includes understanding document contents \cite{vdoc,formnlu}, the document layout structures \cite{docparser} and the recognition of document elements \cite{luo2022doc,layoutparser}.

The existing document VQA mainly examines the understanding of the document in terms of contextual understanding \cite{mathew2021docvqa,tanaka2021visualmrc} and key information extraction \cite{sroie,cord}. Their questions are designed to ask about certain contents on a document page. For example, the question ``What is the income value of consulting fees in 1979?" expects the specific value from the document contents. Such questions examine the model's ability to understand questions and document textual contents simultaneously. 

Apart from the contents, the other important aspect of a document is its structured layout which forms the content hierarchically. Including such structural layout understandings in the document, the VQA task is also critical to improve the model's capabilities in understanding the documents from a high level. Because in real-world document understandings, apart from querying about certain contents, it is common to query a document from a higher level. For example, a common question would be ``What is the figure on this page about?" and answering such a question requires the model to recognize the figure element and understand that the figure caption, which is structurally associated with the figure, should be extracted and returned as the best answer. 

Additionally, the existing document VQA limits the scale of document understanding to a single independent document page \cite{mathew2021docvqa,tanaka2021visualmrc}. But most document files of human's daily work are multi-page documents with successively logical connections between pages. It is a more natural demand to holistically understand the full document file and capture the connections of textual contents and their structural relationships across multiple pages rather than the independent understanding of each page. Thus, it is significant to expand the current scale of page-level document understanding to the full document-level.

In this work, we propose a new document VQA dataset, PDF-VQA, that contains questions to comprehensively examine document understandings from the aspects of 1)document element recognition 2) and their structural relationship understanding 3) from both page-level and full document-level. Specifically, we set up three tasks for our dataset with questions that target different aspects of document understanding. The first task mainly aims at the document elements recognition and their relative positional relationship understandings on the page-level, the second task focuses on the structural understanding and information extraction on the page level, and the third task targets the hierarchical understanding of document contents on the full document level. Moreover, we adopted the automatic question-answer generation process to save human annotation time and enrich the dataset with diverse question patterns. We have also explicitly annotated the relative hierarchical and positional relationships between document elements. As shown in Table \ref{tab:dataset}, our PDF-VQA provides the hierarchically logical relational graph and spatial relational graph, indicating the different relationship types between document elements. This graph information can be used in model construction to learn the document element relationships. We also propose a graph-based model to give insights into how those graphs can be used to gain a deeper understanding of document element relationships from different aspects. 

Our contributions are summarized as 1) We propose a new document-based VQA dataset to examine the document understanding of comprehensive aspects, including the document element recognition and the structural layout understanding; 2) We are the first to boost the scale of document VQA questions from the page-level to the full document level; 3) We provide the explicit annotations of spatial and hierarchically logical relation graphs of document elements for the easier usage of relationship features for future works; 4) We propose a strong baseline for PDF-VQA by adopting the graph-based components.

\begin{table}[t]
\caption{Summary of conventional document-based VQA. Answer type abbreviations are MCQ: Multiple Choice; Ex: Extractive; Num: Numerical answer; Y/N: yes/no; Ab: Abstractive. Datasets with a tick mark in Text Info. the column provides the textual information/OCR tokens on the image/document page ROI. LR graph: logical relational graph; SR graph: spatial relational graph.} \label{tab:dataset}
    \begin{center}
    \begin{adjustbox}{max width=\textwidth}
    \begin{tabular}{l|l|l|l|l|l|c|c}
    \hline
    \textbf{Dataset} & \textbf{Source} & \textbf{Q. Coverage} &\textbf{Answer Type} & \textbf{Img. \#} & \textbf{Q. \#} & \textbf{Text Info.} & \textbf{Relation Info.}\\ 
    \hline
    TQA \cite{kembhavi2017you} & Science Diagrams & diagram contents & MCQ & 1K & 26K & \cmark &  \xmark\\
    DVQA \cite{kafle2018dvqa} & Bar charts & chart contents & Ex, Num, Y/N & 300K & 3.4M & \cmark  &  \xmark\\
    FigureQA \cite{kahou2017figureqa} & Charts & chart contents & Y/N & 180K & 2.4M & \xmark & \xmark\\
    PlotQA \cite{methani2020plotqa} & Charts & chart contents & Ex, Num, Y/N & 224K & 29M & \cmark & \xmark\\
    LEAFQA \cite{chaudhry2020leaf} & Charts & chart contents & Ex, Num, Y/N & 250K & 2M & \xmark & \xmark\\
    DocVQA \cite{mathew2021docvqa} & Single Doc Page & doc contents & Ex & 12K & 50K & \cmark & \xmark\\
    VisualMRC \cite{tanaka2021visualmrc} & Webpage Screenshot & page contents & Ab & 10K & 30K & \cmark & \xmark\\
    InfographicVQA \cite{mathew2022infographicvqa} & Infographic & graph contents & Ex, Num & 5.4K & 30K & \cmark & \xmark\\
    \hline
    \textbf{PDF-VQA TaskA} & Single Doc Page & doc elements & Ex, Num, Y/N & 12k & 81K & \cmark &  \multirow{3}{*}{\makecell{LR graph\\ SR graph}}\\
    \textbf{PDF-VQA TaskB} & Single Doc Page & doc structure & Ex & 12K & 54K & \cmark  & \\
    \textbf{PDF-VQA TaskC} & Entire Doc & doc contents & Ex & 1147 & 5.7K & \cmark  & \\
    \hline
    \end{tabular}
    \end{adjustbox}
    \end{center}
\end{table}

\section{Related Work}
Since the VQA task was introduced \cite{antol2015vqa}, the image source of the VQA task could be divided into three types: realistic/synthetic photos, scientific charts, and document pages. \textbf{\emph{VQA with realistic or synthetic photos}} is widely known as the conventional VQA \cite{antol2015vqa,goyal2017making,johnson2017clevr,hudson2019gqa}. These realistic photos contain diverse object types and the questions of the conventional VQA query about the recognition of objects and their attributes and the positional relationship of the objects. The later proposed scene text VQA problem \cite{mishra2019ocr,singh2019towards,biten2019scene,wang2020general} involves realistic photos with scene texts, such as the picture of a restaurant with its brand name. The questions of scene text VQA query about recognising the scene texts associated with objects in the photos. \textbf{\emph{VQA with scientific charts}} \cite{kafle2018dvqa,kahou2017figureqa,chaudhry2020leaf,methani2020plotqa} contain the scientific-style plots, such as bar charts. The questions usually query trend recognition, value comparison, and the identification of chart properties. \textbf{\emph{VQA with document pages}} involves images of various document types. For example, the screenshots of web pages that contain short paragraphs and diagrams \cite{tanaka2021visualmrc}, info-graphics \cite{mathew2022infographicvqa}, and single document pages of scanned letters/reports/forms/invoices \cite{mathew2021docvqa}. These questions usually query the textual contents of a document page, and most answers are text spans extracted from the document pages.

VQA tasks on document pages are related to Machine Reading Comprehension (MRC) tasks in terms of questions about the textual contents and answered by extractive text spans. Some research works \cite{mathew2021docvqa,tanaka2021visualmrc} also consider it as an MRC task, so it can be solved by applying language models on the texts extracted from the document pages. However, input usage is the main difference between MRC and VQA. Whereas MRC is based on pure texts of paragraphs and questions, document-based VQA focuses on the processing of image inputs and questions. Our PDF-VQA is based on the document pages of published scientific articles, which requires the simultaneous processing of PDF images and questions. We compare VQA datasets of different attributes in Table~\ref{tab:dataset}. While the questions of previous datasets mainly ask about the specific contents of document pages or the certain values of scientific charts/diagrams, our PDF-VQA dataset questions also query the document layout structures and examine the positional and hierarchical relationships understandings among the recognized document elements.

\begin{table}[t]
\caption{Data Statistics of Task A, B, and C. The numbers in \textit{Image} row for Task A/B refer to the number of document pages but the entire document number for Task C.}\label{tab:stats_table}
    \begin{center}
    \begin{adjustbox}{max width = 0.65\linewidth}
    \begin{tabular}{p{0.1\textwidth}p{0.15\textwidth}p{0.1\textwidth}p{0.1\textwidth}p{0.1\textwidth}p{0.1\textwidth}}
    \hline
    \textbf{Task} & \textbf{Type} & \textbf{Train} & \textbf{Valid} & \textbf{Test} & \textbf{Total}\\ 
    \hline
    \multirow{2}{*}{Task A} 
    & Image & 8,593 & 1,280 & 2,464 & 12,337\\ 
    & Question & 59,688 & 7,247 & 14,150 & 81,085 \\
                           \hline
    \multirow{2}{*}{Task B} 
    & Image & 8,593 & 1,280 & 2,464 & 12,337\\ 
    & Question & 37,428 & 5,660 & 10,784 & 53,872 \\
                           \hline
    \multirow{2}{*}{Task C} 
    & Document & 800 & 115 & 232 &1,147\\ 
    & Question & 3,951 & 581 & 1,121 & 5,653\\
                           \hline
    \end{tabular}
    \end{adjustbox}
    \end{center}
    \vspace{-1em}
\end{table}
\vspace*{-0.5em}

\section{PDF-VQA Dataset}\label{sec:PDF-VQA-dataset}

Our PDF-VQA dataset contains three subsets for three different tasks to mainly examine the different aspects of document understanding: Task A) Page-level Document Element Recognition, B) Page-level Document Layout Structure Understanding, and C) Full Document-level Understanding. Detailed dataset statistics are in Table~\ref{tab:stats_table}.

\begin{figure}[t]
 \centering
 \includegraphics[width=0.62\linewidth]{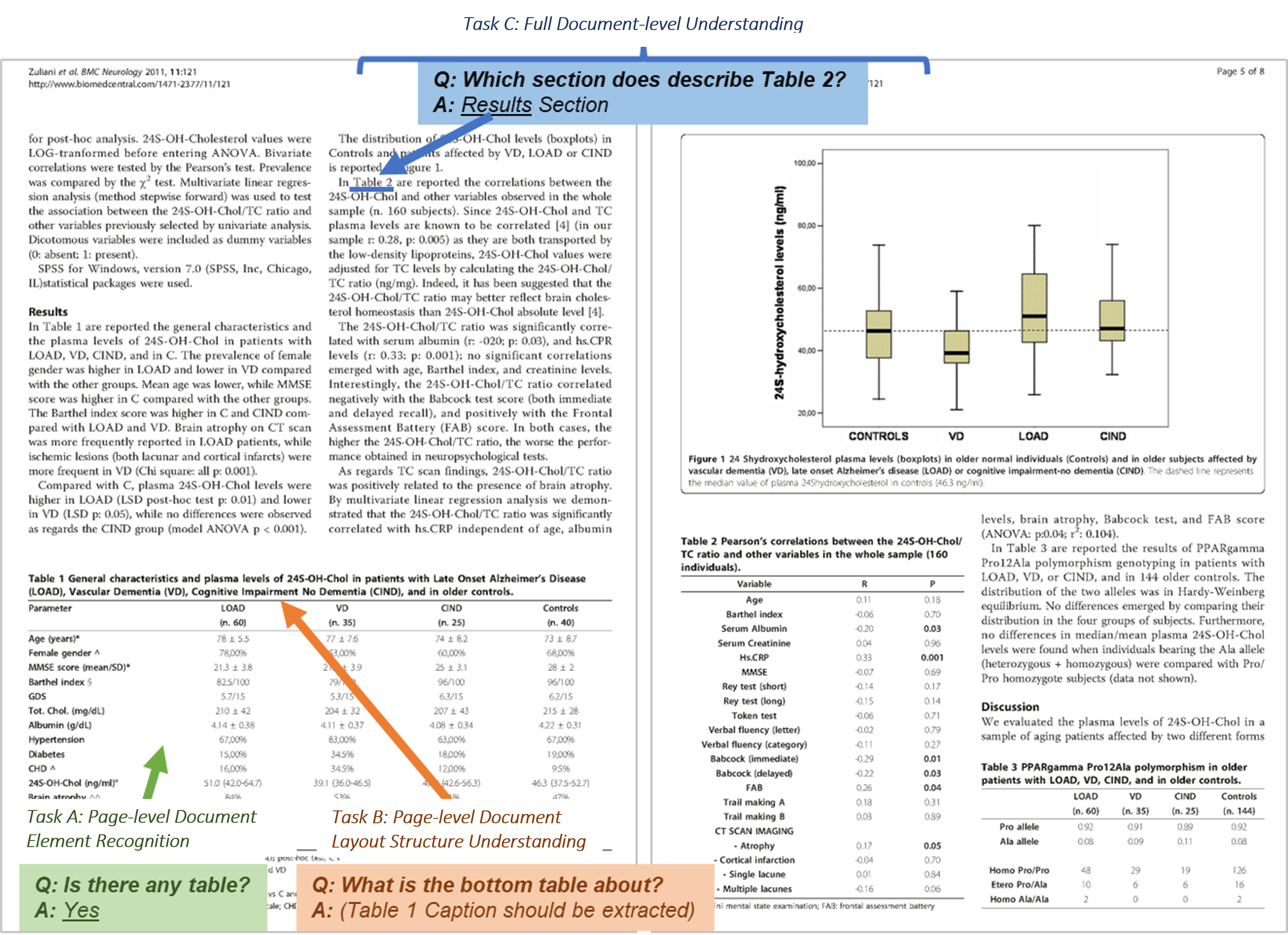}
 \caption{PDF-VQA sample questions and document pages for Task A, B, and C.}
 \label{fig:system}
\end{figure}

\textbf{Task A} aims to examine the document element recognition and their relative spatial relationship understanding on the document page level. Questions are designed into two types to verify the existence of the document elements and count the element numbers. Both question types examine relative spatial relationships and understandings between different document elements. For example, ``Is there any table \textit{below} the 'Results' section?" in Figure~\ref{fig:system} and "How many tables are on this page?". Answers are yes/no and numbers from a fixed answer space.

\textbf{Task B} focuses on understanding the document layout structures spatially and logically based on the recognized document elements on the document page level and extracting the relevant texts as answers to the questions. There are two main question types: structural understanding and object recognition. The structural understanding questions relate to examining spatial structures from both relative positions or human reading order. For example, ``What is the \textit{bottom} section about?" requires understanding the document layout structures from the relative bottom position and ``What is the \textit{last} section about?" requires identifying the last section based on the human reading order of a document. The object recognition questions explicitly contain a specific document element in the questions and require to recognition of the queried element first, such as the question ``What is the bottom table about?" in Figure~\ref{fig:system}. Answering these two types of questions require a logical understanding of the hierarchical relationships of document elements. For instance, based on the textual contents, the section title would be a logically high-level summarization of its following section and is regarded as the answer to ``What is the last section about?". Similarly, a table caption is logically associated with a table; table caption contents would best describe a table. 

\textbf{Task C} questions have a sequence of answers extracted from multi-pages of the full document. It enhances the document understanding from the page to the full document level. Answering a question in Task C requires reviewing the full document contents and identifying the contents hierarchically related to the queried item in the question. For example, the question ``Which section does describe Table 2?" in Figure~\ref{fig:system} requires the identification of all the sections of the full document that have described the queried table. The answers to such questions are the texts of the corresponding section titles extracted as the high-level summarization of the identified sections. Identifying the items at the higher-level hierarchy of the queried item is defined as the parent relation understanding the question in PDF-VQA. Oppositely, Task C also contains the questions of identifying the items at the lower-level hierarchy of the queried item, and such questions are defined as the child relation understanding. For example, a question, ``What does the `Methods' section about?" requires extracting all the subsection titles as the answer. 

The detailed question type distribution of each task is shown in Table~\ref{tab:qtype_table}. 

\subsection{Data Source}
Our PDF-VQA dataset collected the PDF version of visually-rich documents from the PubMed Central (PMC) Open Access Subset\footnote{\url{https://www.ncbi.nlm.nih.gov/pmc/tools/openftlist/}}. Each document file has a corresponding XML file that provides the structured representations of textual contents and graphical components of the article\footnote{It follows the XML schema module provided by the Journal Archiving and Interchange Tag Suite created by the National Library of Medicine (NLM) \url{https://dtd.nlm.nih.gov/}}. We applied the pretrained Mask-RCNN~\cite{zhong2019publaynet} over the collected document pages to get the bounding boxes and categories for each document element. The categories initially consisted of five common PDF document element types: \textit{title}, \textit{text}, \textit{list}, \textit{figure}, and \textit{table}. We then labelled the \textit{text} elements that are positionally closest to the tables and figures into two additional categories \textit{table caption} and \textit{figure caption} respectively.

\subsection{Relational Graphs}\label{subsec:parent-child graph}
Visually rich documents of scientific articles consist of fixed layout structures and hierarchically logical relationships among the sections, subsections and other elements such as tables/figures and table/figure captions. Understanding such layout structures and relationships is essential to boost the understanding of this type of document. 
The graph has been used as an effective method to represent the relationships between objects in many tasks \cite{davis2021visual,zhang2021vsr,zhang2022multimodal,luo2022doc}. Inspired by this, for each document, we annotated the hierarchically \textit{logical relational} graph (LR graph) and \textit{spatial relational} graph (SR graph) to explicitly represent the logical and spatial relationships between document elements respectively. Those two graphs can be directly used by any deep-learning mechanisms to enhance the feature representation. In Section~\ref{sec:our_model}, we propose a graph-based model to enlighten how such relational information can solve the PDF-VQA questions.
The SR graph indicates the relative spatial relationships between document elements based on their absolute geometric positions with their bounding box coordinates. For each document element of a single document page, we identify its relative spatial relationships with all the other document elements among eight spatial types: \textit{top, bottom, left, right, top-left, top-right, bottom-left} and \textit{bottom-right}. The LR graph indicates the potential affiliation between document elements by identifying the parent object and their children's objects based on the hierarchical structures of document layouts. We follow \cite{luo2022doc} to annotate the parent-child relations between the document elements in a single document page to generate the LR graph. The graph of the full document of multiple pages are augmented by the graphs of its document pages.

\begin{table*}[t]
\caption{Ratio and exact number of various question types of Task A, B and C.}\label{tab:qtype_table}
    \begin{center}
    \begin{adjustbox}{width = 0.85\linewidth}
    \begin{tabular}{p{0.2\textwidth}p{0.5\textwidth}p{0.2\textwidth}p{0.2\textwidth}}
    \hline
    \textbf{Tasks} & \textbf{Question Type} & \textbf{Percentage} & \textbf{Total}\\ 
    \hline
    \multirow{2}{*}{Task A} 
    & Counting & 17.74 & 14,387\\ 
    & Existence & 82.26 & 66,698 \\
                           \hline
    \multirow{2}{*}{Task B} 
    & Structural Understanding & 88.58 & 47,722\\ 
    & Object Recognition & 11.42 & 6,150 \\
                           \hline
    \multirow{2}{*}{Task C} 
    & Parent Relationship Understanding &  79.71 &4,506\\ 
    & Child Relationship Understanding &  20.29 & 1,147\\
                           \hline
    \end{tabular}
    \end{adjustbox}
    \end{center}

\end{table*}

\subsection{Question Generation}

Visually rich documents of scientific articles have consistent spatial and logical structures. The associated XML files of these documents provide detailed logical structures between semantic entities. Based on this structural information and the pre-defined question template, we applied an automatic question-generation process to generate large-scale question-answer pairs efficiently. For example, the question \textit{``How many tables are above the `Discussion'?"} is generated from the question template \textit{``How many \textlangle E1\textrangle~are \textlangle R\textrangle~the `\textlangle E2\textrangle'?"} by filling the masked terms \textit{\textlangle E1\textrangle}, \textit{\textlangle R\textrangle} and \textit{\textlangle E2\textrangle} with document element label (``table"), positional relationship (``above") and title name extracted from document contents (``Discussion") respectively. We prepare each question template with various language patterns to diversify the questions. For instance, the above template can also be written as \textit{``What is the number of \textlangle E1\textrangle~are \textlangle R\textrangle~the `\textlangle E2\textrangle'?"}. We have 36, 15, and 15 question patterns for Task A, B, and C, respectively. We limit the parameter values of the document element label to only \textit{title, list, table, figure} as asking for the number/existence/position of \textit{text} elements would be less valuable. The parameter values include four document element labels, eight positional relationships (\textit{top, bottom, left, right, top-left, top-right, bottom-left} and \textit{bottom-right}), ordinal form (\textit{first, last}) and the texts from document contents (e.g. section title, references, etc.). We also replace some parameter values with their synonyms, such as \textit{``on the top of"} for \textit{``above"}.

To automatically generate the ground truth answers to our questions, we first represent each document page (for Task A and B)/the full document (for Task C) with all the document elements and the associated relations from the two relational graphs as in Section~\ref{subsec:parent-child graph}. We then apply the functional program, which is uniquely associated with each question template and contains a sequence of functions representing a reasoning step, over such document(page) representations to reach the answer. For example, the functional program for question \textit{``How many tables are above of the `Discussion'?"} consists of a sequence of functions $\textit{filter-unique}\rightarrow\textit{query-position}\rightarrow\textit{filter-category}\rightarrow\textit{count}$ to filter out the document elements that satisfy the asked positional relationships and count the numbers of them as the ground-truth answer. 

Moreover, we conduct the question balancing from answer-based and question-based aspects to avoid question-conditional biases and balance the answer distributions. Firstly, we conduct an answer-based balancing by down-sampling questions based on the answer distribution. We identify the QA pairs with large ratios, divide identified questions into groups based on the patterns, and reduce QA pairs with large ratios until the answer distributions are balanced. After that, we further conducted the question-based balancing to avoid duplicated question types. To achieve this, we smooth over the distributions of parameter values filled in the question templates by removing the questions with large proportions of certain parameter values until the balanced distribution of parameter value combinations. Since the parameter values of Task C question templates are almost unique, as all of them are the texts from document contents, we did not conduct the balancing over Task C. After the balancing, Task A questions are down-sampled from 444,967 to 81,085, and Task B questions are down-sampled from 246,740 to 53,872.

\section{Dataset Analysis and Evaluation} 
\subsection{Dataset Analysis}
\begin{figure}[t]
 \centering
 \includegraphics[width=0.9\linewidth]{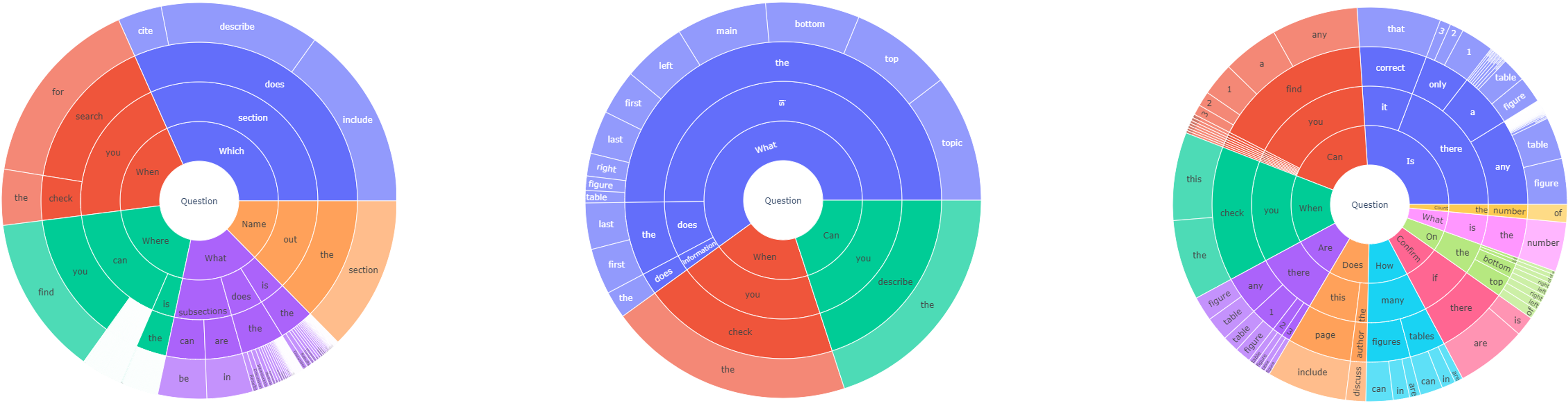}
 \caption{The top 4 words of questions in Task A, B and C.}
 \label{fig:top4}
 \vspace{-0.5em}
\end{figure}

\begin{figure}[ht]
 \centering
 \includegraphics[width=1\linewidth]{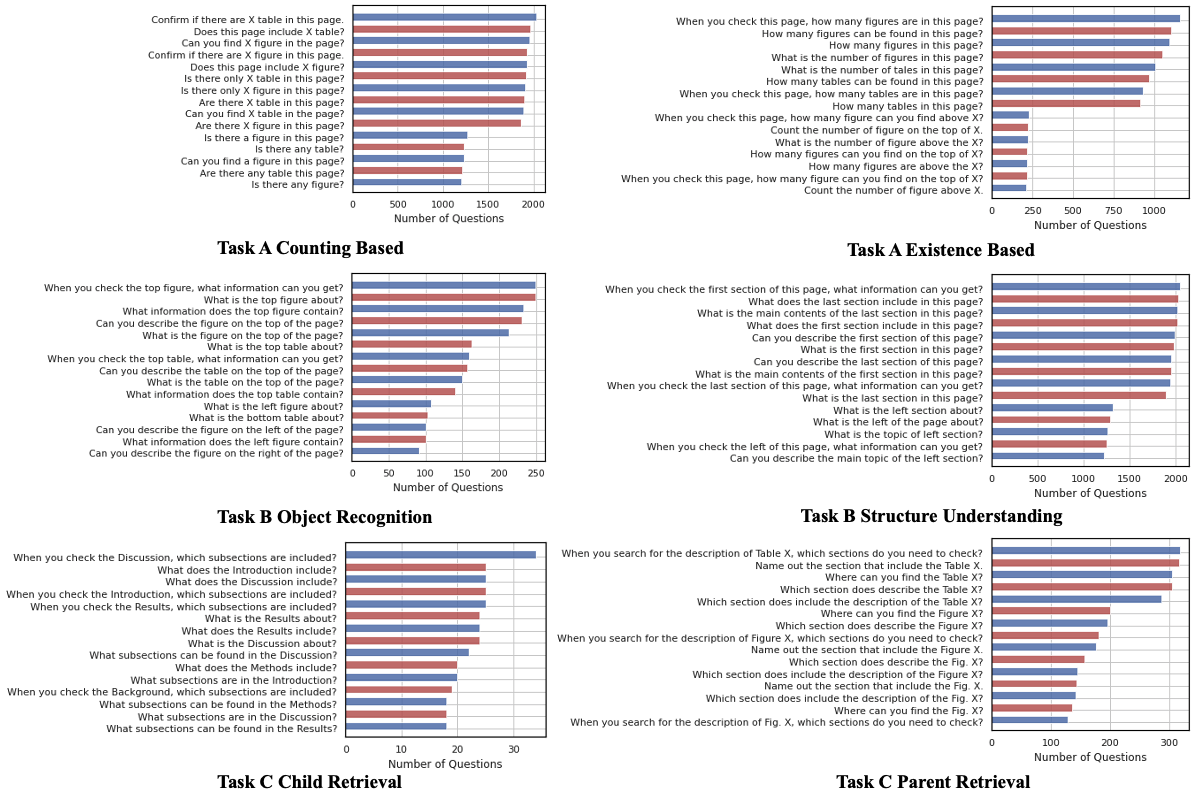}
 \caption{Top 15 Frequency Questions of Task A, B and C.}
 \label{fig:top15}
\end{figure}

The average number of questions per document page/document in Task A, B, and C are 6.57, 4.37, and 4.93. The average question length for Task A, B and C are 25, 10 and 15, respectively \footnote{We provide the question length distribution analysis in Appendix C.1}. A sunburst plot showing each task's top 4 question words is shown in Figure~\ref{fig:top4}. We can see that Task A question priors are more diverse to complement the simplicity of document element and position recognition questions and to prevent the model from memorizing question patterns. For Task B and C, question priors distribute over ``What", ``When", ``Can you", ``Which". And we also specifically design questions in a declarative sentence with ``Name out the section..." in Task C. 13.43\%, 0.24\% and 29.38\% of the questions in Task A, B, and C are unique questions. This unique question ratio seems low compared to other document-based VQA datasets. This is because, rather than only aiming at the textual understanding of certain page contents, our PDF dataset targets more the spatial and hierarchically structural understandings of document layouts. Our questions are generally formed to ask about the document structures from a higher level and thus contain less unique texts that are associated with the specific contents of each document page. Answers for Task A questions are from the fixed answer space that contains eight possible answers: \textit{``yes", ``no", ``0", ``1", ``2", ``3", ``4"} and \textit{``5"}. Answers for Task B and C are texts retrieved from the document page/entire document. We also analyzed the top 15 frequent question patterns in Task A, B and C as shown in Figure~\ref{fig:top15} to show the common questions of each question type in each task. We used a placeholder ``X'' to replace the different figures, table numbers or section titles that would exist in the questions to present the common question patterns in this analysis.

\subsection{Human Evaluation}

\begin{table}[htb]
\caption{Positive rates (\textbf{Pos(\%)}) and Fleiss Kappa Agreement (\textbf{Kappa}) of human evaluation.}\label{tab:human_evaluation}
    \begin{center}
    \begin{adjustbox}{max width = 0.7\linewidth}
    \begin{tabular}{p{0.2\textwidth}|p{0.1\textwidth}p{0.1\textwidth}|p{0.1\textwidth}p{0.1\textwidth}|p{0.1\textwidth}p{0.1\textwidth}}
    \hline
    \multicolumn{1}{c}{} & \multicolumn{2}{c}{\textbf{Task A}} & \multicolumn{2}{c}{\textbf{Task B}} & \multicolumn{2}{c}{\textbf{Task C}}\\
    \hline
    \textbf{Perspective}& \textbf{Pos(\%)} & \textbf{Kappa} & \textbf{Pos(\%)} & \textbf{Kappa} & \textbf{Pos(\%)} & \textbf{Kappa} \\
    \hline
    Relevance&  98.46 & 94.02 & 91.67 & 77.07 & 100 & 100\\
    Correctness&  99.49 & 98.12 & 89.44 & 72.56 & 94.55 & 80.93\\
    Meaningfulness&  96.94 & 88.97 & 93.61 & 77.67 &99.27 &97.34\\
    \hline
    \end{tabular}
    \end{adjustbox}
    \end{center}
    \vspace{-1em}
\end{table}
To evaluate the quality of automatically generated question-answer pairs, we invited ten raters, including deep-learning researchers and crowd-sourcing workers. Firstly, to determine the relevance between the question and the corresponding page/document, we define the \textit{Relevance} criteria. Correspondingly, we define \textit{Correctness} to determine whether the auto-generated answer is correct to the question. In addition, we ask raters to judge whether our QA pairs are meaningful and possibly appear in the real world by using \textit{Meaningfulness} criteria \footnote{More details and human evaluation survey examples can be found in Appendix B.}. After we collect the raters' feedback, we calculate the positive rate of each perspective and apply Fleiss Kappa to measure the agreements between multiple raters, as can be seen in Table~\ref{tab:human_evaluation}. All three tasks achieve decent positive rates with substantial or almost perfect agreements. For Task A, \textit{Relevance} and \textit{Correctness} can reach positive rates with nearly perfect agreements. Few raters gave negative responses regarding the \textit{Meaningfulness} of questions about the existence of tables or figures, while those questions are crucial to understanding the document layout for any upcoming table/figure contents understanding questions. In Task B, all three perspectives achieve high positive rates with substantial agreements. The disagreements about Task B mainly come from the questions with no specific answer (N/A), some raters thought those questions were incorrect and meaningless, but these questions are crucial to understanding the commonly appearing real-world cases. Because it is possible that a page does not contain the queried elements in the question, and no specific answer is a reasonable answer for such cases. Finally, for Task C, both positive rates and agreement across three perspectives are notable. In addition, except for three perspectives, raters agree most of the questions in Task C need cross-page understanding (the positive rate is 82.91\%).

\section{Baseline Models} 
We experimented with several baselines on our PDF-VQA dataset to provide a preliminary view of different models' performances. We choose the vision-and-language models that have proved good performances on VQA tasks and a language model as listed in Table~\ref{tab:performance}. We followed the original settings of each baseline but only made modifications on the output layers to suit different PDF-VQA tasks\footnote{The detailed baseline model setup can be found in Appendix C, and the code for the baseline model and our proposed model will be released in GitHub after paper acceptance.}.

\section{Proposed Model: LoSpa}\label{sec:our_model}
In this paper, we introduce a strong baseline, Logical and Spatial Graph-based model (\textbf{\textit{LoSpa}}), which utilizes logical and spatial relational information based on logical (LR) and spatial (SR) graphs introduced in Section~\ref{subsec:parent-child graph}.

\begin{figure*}[t]
 \centering
 \includegraphics[width=0.7\linewidth]{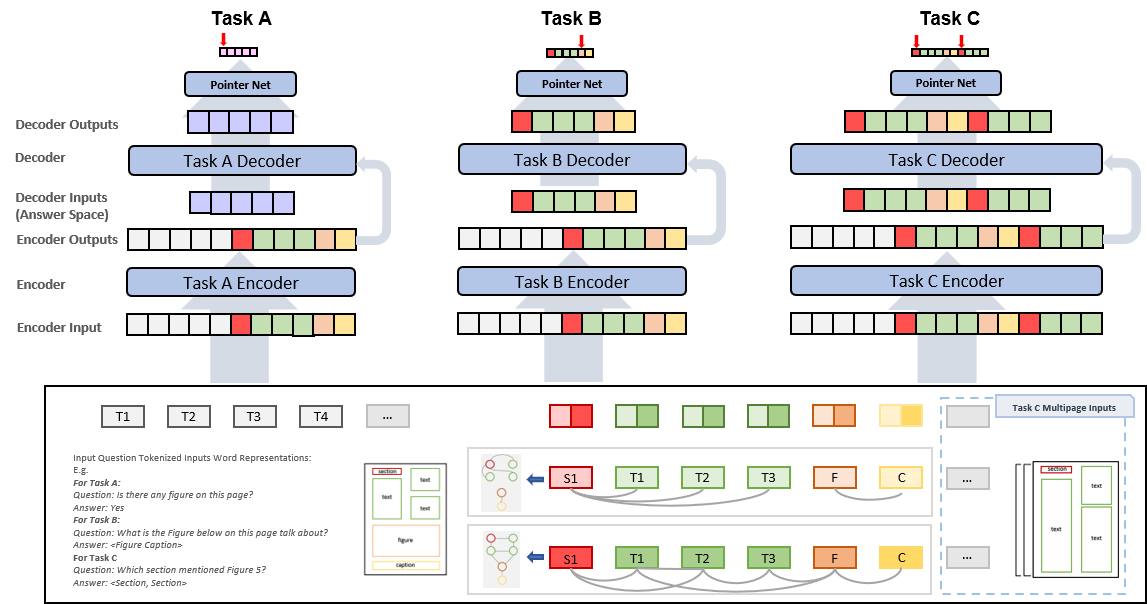}
 \caption{Logical and Spatial Graph-based Model Architecture for three tasks. Task A, B and C use the same relational information to enhance the object representation but different model architectures in the decoding stage. }
 \label{fig:model_architecture}
\end{figure*}

\textbf{Input Representation}: we treat questions as sequential plain text inputs and encode them by BERT. For document elements of given document page $I$ such as \textit{Title}, \textit{Text}, \textit{Figure}, we use pre-trained ResNet-101 backbones to extract visual representations $X_v \in \mathbb{R}^{N\times d_f}$ and use [CLS] token from BERT as the semantic representation $X_s \in \mathbb{R}^{N\times d_s}$ for the texts of each document element.

\textbf{Relational Information Learning}: 
We construct two graphs: logical graph $\mathcal{G}_{l} = \left ( \mathcal{V}_{l}, \mathcal{E}_{l} \right )$ and spatial graph $\mathcal{G}_{s} = \left ( \mathcal{V}_{s}, \mathcal{E}_{s} \right )$ for each document page. For the logical graph $\mathcal{G}_{l}$, based on \cite{luo2022doc}, we define the semantic feature as node representation $\mathcal{V}_{l}$ and the existence of parent-child relation between document elements (extracted from the logical relational graph annotation in our dataset) as binary edge values $\mathcal{E}_{l}$ $\left \{0, 1  \right \}$. Similarly, for spatial graph $\mathcal{G}_{s}$, we follow \cite{luo2022doc} to use the visual features of document elements as node representation $\mathcal{V}_{s}$ and the distance with two nearest document elements to weight edge value $\mathcal{E}_{s}$.

For each document page $I$, we take $X_s \in \mathbb{R}^{N\times d_s}$ and $X_v \in \mathbb{R}^{N\times d_f}$ as the initial node feature matrix for $\mathcal{G}_{l}$ and $\mathcal{G}_{s}$ respectively. These initial node features are fed into a two-layer Graph Convolution Network (GCN) and trained by predicting each node category. After the GCN training, we extract the first layer hidden states as the updated node representations $X_s' \in \mathbb{R}^{N\times d}$ and $X_v' \in \mathbb{R}^{N\times d}$ that has augmented the relational information between document elements for $\mathcal{G}_{s}$ and $\mathcal{G}_{f}$ respectively, where $d=768$. For each aspect feature, we conduct separated linear transformations to the initial feature matrices ($X_v$/$X_s$) and the updated feature matrices ($X_s'$/$X_v'$). Inspired by \cite{luo2022doc}, we apply the element-wise max-pooling over them. The pooled features $X''_s$ and $X''_v$ are the final semantic and visual representations of nodes enhanced by logical and spatial relations, respectively. Finally, we concatenate semantic and visual features of each document element, yielding relational information enriched multi-modal object representations $O_1, O_2,..., O_N$.

\textbf{QA prediction}: We sum up the object features $O_1, O_2,..., O_N$  with positional embedding to integrate the information of document elements orders, which are inputs into multiple transformer encoder layers together with the results of the sequence of question word features $q_1, q_2,...,q_T$. We pass the encoder outputs into the transformer decoders and apply a pointer network upon the decoder output to predict the answers. We apply a one-step decoding process each time using the word embedding $w_{i}$ of one answer from the fixed answer space as the decoder input. Let the $z_{i}^{dec}$ be the decoder output for the decoder input $w_{i}$; we then conduct the score $y_{t,i}$ between $z_{i}^{dec}$ and the answer word embedding $w_{i}$ following $y_{t,i} = \left ( w_{i} \right )^{T}z_{t}^{dec} + b_{i}^{dec}$, where $i = 1,..., C$, and $C$ are the total answer numbers of the fixed answer space for each task. We apply a softmax function over all the scores $y_{1},...,y_{C}$ and choose the answer word with the highest probability as the final answer for the current image-question pair. We treat Task B and C as the same classification problem as Task A, where the answers are fixed to 25 document element index numbers for Task B and 400 document element index numbers for Task C. The index numbers for document elements start from 0 and increase following the human-reading order (i.e. top to bottom, left to right) over a single document page (for Task B) and across multiple document pages (for Task C). OCR tokens are extracted from the document element with the corresponding predicted index number for the final retrieved answers for Task B and C questions. We use the Sigmoid function for Task C questions with multiple answers and select all the document elements whose probability has passed 0.5.

\section{Experiments}
\subsection{Performance Comparison}
We compare the performances of baseline models and our proposed relational information-enhanced model over three tasks of our PDF-VQA dataset in Table~\ref{tab:performance}. All the models process the questions in the same way as the sequence of question words encoded by pretrained BERT but differ in other features' processing. The three large vision-and-language pretrained models (VLPMs): VisualBERT, ViLT and LXMERT, achieved better performances than other baselines with inputting only question and visual features. The better performance of VisualBERT than ViLT indicates that object-level visual features are more effective than image patch representations on the PDF-VQA images with segmented document elements. Among these three models, LXMERT, which used the same object-level visual features and the additional bounding box features, achieved the best results over Task A and B, indicating the effectiveness of bounding box information in the cases of PDF-VQA task. However, its performance on Task C is lower than VisualBERT. This might be because Task C inputs the sequence of objects (document elements) from multiple pages. The bounding box coordinates are independent on each page and therefore cause noise during training. Surprisingly, LayoutLM2, pretrained on document understanding datasets, achieved much lower accuracy than the three VLPMs. This might be because LayoutLM2 used token-level visual and bounding box features, which are ineffective for the whole document element identification. Compared to LayoutLM2 used the token-level contextual features, M4C, as a non-pretrained model, inputting object-level bounding box, visual and contextual features achieved higher performances. Such results further indicate that the object-level features are more effective for our PDF-VQA tasks. The object-level contextual features of each document element are represented as the [CLS] hidden states from the pretrained BERT model inputting the OCR token sequence extracted from each document element. 

Our proposed \textit{LoSpa} achieves the highest performance compared to all baselines, demonstrating the effectiveness of our adopted GCN-encoded relational features. Overall, all models' performances are the highest on Task A among all tasks due to the relatively simple questions associated with object recognition and counting. The performances of all the models naturally dropped on Task B when the ability of contextual and structural understanding are simultaneously required. Performances on Task C are the lowest for all models. It indicates the difficulty of document-level questions and produces massive room for improvement for future research on this task.

\begin{table}[t!]
\caption{Performance Comparison over Task A, B, and C. Acronym of feature aspects: Q: Question features; B: Bounding box coordinates; V: Visual appearance features; C: Contextual features; R: Relational Information.} \label{tab:performance} 
    \begin{center}
    \begin{adjustbox}{max width=0.9\linewidth}
    \begin{tabular}{p{0.2\textwidth}|p{0.06\textwidth}p{0.06\textwidth}p{0.06\textwidth}p{0.06\textwidth}p{0.06\textwidth}|p{0.1\textwidth}p{0.1\textwidth}|p{0.1\textwidth}p{0.1\textwidth}|p{0.1\textwidth}p{0.1\textwidth}}
    \hline
    \multicolumn{1}{c}{} & \multicolumn{5}{c}{\textbf{Feature Aspects}} & \multicolumn{2}{c}{\textbf{Task A}} & \multicolumn{2}{c}{\textbf{Task B}} & \multicolumn{2}{c}{\textbf{Task C}}\\
    \hline
    \textbf{Model} & \textbf{Q.} & \textbf{B.} &\textbf{V.} & \textbf{C.} & \textbf{R.} & \textbf{Val.} & \textbf{Test} & \textbf{Val.} & \textbf{Test} & \textbf{Val.} & \textbf{Test} \\
    \hline
    VisualBERT \cite{li2019visualbert} 
    & \cmark & \xmark & \cmark & \xmark & \xmark & 92.72 & 92.34 & 82.00 & 79.43 & 21.55 & 18.52 \\
    ViLT \cite{kim2021vilt} 
    & \cmark & \xmark & \cmark & \xmark & \xmark & 90.82 & 91.31 & 54.36 & 53.45 & 10.21 & 9.87 \\
    LXMERT \cite{tan2019lxmert} 
    & \cmark & \cmark & \cmark & \xmark & \xmark & 94.34 & 94.41 & 86.61 & 86.36 & 16.37 & 14.41 \\
    BERT \cite{devlin2019bert}
    & \cmark & \xmark & \xmark & \cmark & \xmark & 82.35 & 81.87 & 22.41 & 23.64 & - & - \\
    LayoutLM2 \cite{xu2021layoutlmv2}
    & \cmark & \cmark & \cmark & \cmark & \xmark & 83.27 & 83.49 & 22.70 & 23.73 & - & - \\
    M4C \cite{hu2020iterative} 
    & \cmark & \cmark & \cmark & \cmark & \xmark & 87.89 & 87.98 & 56.80 & 55.29 & 12.14 & 13.77 \\
    \hline
    \textbf{Our LoSpa} & \cmark & \cmark & \cmark & \cmark & \cmark & \textbf{94.98} & \textbf{94.55} & \textbf{91.10} & \textbf{90.64} & \textbf{30.21} & \textbf{28.99} \\
    \hline
    \end{tabular}
    \end{adjustbox}
    \end{center}
    \vspace*{-1em}
\end{table}
\hspace{-1em}

\subsection{Relational Information Validation}
To further demonstrate the influences of relational information on document VQA tasks, we perform the ablation studies on each task, as shown in Table~\ref{tab:ablation_study}. For all three tasks, adding both aspects of relational information can effectively improve the performance of our \textit{LoSpa} model. Firstly, Spatial relation (SR) enhanced models can make the models of all three tasks more robust. Regarding logical relation (LR), it can lead to more apparent improvements on Task B since Task B involves more questions that require understanding document structure more comprehensively. Moreover, since the graph representation of two relation features is trained on the training set, most of the test set performance is lower than the validation set during the QA prediction stage.

\begin{table}[t]
\caption{Validating the effectiveness of proposed logical-relation (LR) and spatial-relation (SR) based graphs.} \label{tab:ablation_study}
    \begin{center}
    \begin{adjustbox}{max width = 0.7\linewidth}
    \begin{tabular}{p{0.3\textwidth}|p{0.1\textwidth}p{0.1\textwidth}|p{0.1\textwidth}p{0.1\textwidth}|p{0.1\textwidth}p{0.1\textwidth}}
    \hline
    \multirow{2}{*}{\textbf{Configurations}} & \multicolumn{2}{c}{\textbf{Task A}} & \multicolumn{2}{c}{\textbf{Task B}} & \multicolumn{2}{c}{\textbf{Task C}}\\
    \cline{2-7}
    & \textbf{Val.} & \textbf{Test} & \textbf{Val.} & \textbf{Test} & \textbf{Val.} & \textbf{Test} \\
    \hline
    \bf None & 94.17 & 94.12 & 90.02 & 89.59 & 27.13 & 27.71\\
    \bf Logical Relation (LR) & 94.59 & 93.72 & 90.97 & \bf 90.67 & 29.22 & 27.91 \\
    \bf Spatial Relation (SR) & 94.58 & 94.27 & 90.39 & 90.02 & 28.11 & 27.90 \\
    \hline
    \bf LR\&SR & \bf 94.98 & \bf 94.55 & \bf 91.10 & 90.64 & \bf 30.21 & \bf 28.99 \\
    \hline
    \end{tabular}
    \end{adjustbox}
    \end{center}
    \vspace*{-1em}
\end{table}

\begin{table}[t]
\caption{Task A, B and C performance on different question types. Same as the overall performance shown previously, the metric of Task A/B is F1 and Task C is Accuracy.} \label{tab:bdA}
    \begin{center}
    \begin{adjustbox}{max width = 0.9\linewidth}
    \begin{tabular}{l|cc|cc|cc|cc|cc|cc}
    \hline
    \multirow{3}{*}{\bf Model} & \multicolumn{4}{c|}{\textbf{Task A}} & \multicolumn{4}{c|}{\textbf{Task B}} & \multicolumn{4}{c}{\textbf{Task C}}\\
    \cline{2-13}
    & \multicolumn{2}{c|}{\textbf{Existence}} & \multicolumn{2}{c|}{\textbf{Counting}} & \multicolumn{2}{c|}{\textbf{Struct-UD}} & \multicolumn{2}{c|}{\textbf{Obj-Reg}} & \multicolumn{2}{c|}{\textbf{Parent}} & \multicolumn{2}{c}{\textbf{Child}}\\
    \cline{2-13}
    & \textbf{Val.} & \textbf{Test} & \textbf{Val.} & \textbf{Test} & \textbf{Val.} & \textbf{Test} & \textbf{Val.} & \textbf{Test}  & \textbf{Val.} & \textbf{Test}  & \textbf{Val} & \textbf{Test} \\
    \hline
    VisualBERT \cite{li2019visualbert} & 94.11 & 91.62 &  92.52 & 92.45 & 83.24 & 80.86 & 71.49 & 70.30& 21.55 & 19.91 & 19.64 & 18.52\\
    ViLT \cite{kim2021vilt} & 92.34 & 93.40 & 90.62 & 91.01 & 53.41 & 51.97 & 59.54 & 61.66& 11.04 & 10.21 & 8.75 & 8.79\\
    LXMERT \cite{tan2019lxmert} & 96.02 & 94.59 & 94.10 & 94.38 & 86.65 & 86.86 & 86.46 & 83.15& 26.66 & 23.57 & 8.56 & 9.51\\
    BERT \cite{devlin2019bert} & 86.25 & 86.04 & 81.80 & 81.31  & 30.42 & 30.55 & 21.37 & 22.33 &-&-&-&-\\
    LayoutLM2 \cite{xu2021layoutlmv2} & 87.22 & 85.78 & 82.70 & 83.19 & 33.18 & 31.80 & 21.55 & 22.63&-&-&-&-\\
    M4C \cite{hu2020iterative} & 90.78 & 89.15 & 87.51 & 87.87& 60.74 & 60.29 & 21.29 & 20.39& 13.63 & 14.34 & 12.21 & 9.89 \\
    \hline
    \textbf{Our LoSpa}& \textbf{97.40}  &  \textbf{95.73} & \textbf{94.39} & \textbf{94.63} &\textbf{91.61} & \textbf{91.14} & \textbf{86.66} 
    & \textbf{87.29}& \textbf{33.14} & \textbf{29.87} & \textbf{29.11} & \textbf{28.74} \\
    \hline
    \end{tabular}
    \end{adjustbox}
    \end{center}
    \vspace{-1em}
\end{table}

\subsection{Breakdown Results}
We conduct the breakdown performance comparison over different question types of each task as shown in Table~\ref{tab:bdA}. Generally, all models' performances on Existence/Structural Understanding/Parent Relation Understanding questions are slightly better than Counting/Object Recognition/Child Relation Understanding questions in tasks A, B and C, respectively, due to their larger question numbers when training. Overall, all models' performances are stable on different question types of each task and follow the same performance trend as on all questions in Table~\ref{tab:performance}. However, M4C's performance on Object Recognition is much lower than its performance on the Structural Understanding questions. This indicates that M4C is more powerful in recognising the contexts and identifying the semantic structures between document elements. However, it does not have enough capacity to identify the elements and related semantic elements simultaneously. Also, the LXMERT's performances on Parent Relation Understanding questions are much better than those on Child Relation Understanding questions. This is because answers to parent questions are normally located on the same page as the queried elements. In contrast, answers to child questions are normally distributed over several pages, which is impacted by the independent bounding box coordinates of each page. The stable performances of M4C over the two question types of task C also indicate that using contextual features would eliminate such issues. Our \textit{LoSpa}, incorporating relational information between document elements, achieves stable performances over both question types in Task C.

\section{Conclusion}
We proposed a new document-based VQA dataset to comprehensively examine the document understanding in conditions of natural language questions. In addition to contextual understanding and information retrieval, our dataset questions also specifically emphasize the importance of document structural layout understanding in terms of comprehensive document understanding. This is also the first dataset that introduces document-level questions to boost the document understanding to the full document level rather than being limited to one single page. We enriched our dataset by providing a Logical Relational graph and a Spatial Relational graph to annotate the different relationship types between document elements explicitly. We proved that such graph information integration enables outperforming all the baselines. We hope our PDF-VQA dataset will be a useful resource for the next generation of document-based VQA models with an entire multi-page document-level understanding and a deeper semantic understanding of vision and language.

\section*{Ethical Consideration}
This study was reviewed and approved by the ethics review committee of the authors' institution and conducted in accordance with the principles of the Declaration. Written informed consent was obtained from each participant.

%
%
%
\bibliographystyle{splncs04}
\bibliography{mybibliography}
\appendix
\section{Question Templates with Examples}
All question templates for three tasks are listed in Table~\ref{tab:taska_templates}, \ref{tab:taskb_templates}, \ref{tab:taskc_templates} with the corresponding real question examples. Task A contains 36 question patterns, including 22 Existence type question patterns and 14 Counting type question patterns. For Task B, the Structural Understanding type contains 10 question patterns, and Object Recognition contains 5. Regarding Task C, there are 5 patterns provided for Child Relation Understanding and 10 patterns designed for Parent Relation Understanding, respectively.
\begin{table*}[!ht]
\begin{center}
    \centering
    \begin{adjustbox}{max width=\textwidth}
    \begin{tabular}{p{0.8\textwidth}p{0.8\textwidth}}
    
    \hline
    
    \multicolumn{1}{c}{\textbf{Question Pattern}} &
    \multicolumn{1}{c}{\textbf{Question Example}} \\
    \hline
    \multicolumn{2}{c}{\textbf{Existence Type Question Patterns}}\\
    \hline
        Is there any [E] on the [pos] of this page?  & Is there any table on the top of this page?   \\ 
        Can you find any [E] on the [pos] of this page? & Can you find any figure on the right of this page?  \\ 
        On the [pos] of this page, is there a [E]? & On the left of this page, is there a table?  \\ 
        Is it correct that there is no [E] at the [pos]?  & Is it correct that there is no figure at the bottom?   \\ 
        When you check the [pos] of this page, can you find any [E]? & When you check the right of this page, can you find any table?  \\ \hline
        Are there any [E1] are [R] the [E2]? & Are there any figures upper the 'Competition analysis'?  \\
        Can you find any [E1] [R] the [E2]? & Can you find any table above the 'Balanced networks'?  \\ 
        Is there a [E1] found [R] the [E2]?  & Is there a table found under the 'Competition analysis'?  \\
        Is it correct that there is no [E1] [R] the [E2]? & Is it correct that there is no table upper the 'Discussion'?  \\
        Confirm if there are any [E1] [R] the [E2]? & Confirm if there are any figures upper the 'Result and Discussion'?  \\
        When you check the page, is there any [E1] [R] the [E2]? & When you check the page, is there any table below the 'Results'?  \\ \hline
        Is there any [E]? & Is there any table?  \\
        Are there any [E] on this page? & Are there any figures in this page?  \\
        Is there a [E] in this page? & Is there a table on this page?  \\
        Can you find a [E] on this page? & Can you find a figure on this page? \\ 
        When you check this page, can you find any [E]? & When you check this page, can you find any table?  \\ \hline
        Is there a [E] on this page? & Is there a 'Results' on this page?  \\ 
        Can you find a [E] on this page? & Can you find a 'Discussion' on this page?  \\
        Does this page include a [E]? & Does this page include a 'Conclusion'?  \\ 
        Can [E] be found on this page? & Can 'Abstract' be found on this page?  \\ 
        When you check this page, can you find [E]? & When you check this page, can you find 'Introduction'?  \\ 
        Confirm if there is [E] on this page. & Confirm if there is an 'Abstract' on this page. \\ \hline
        \multicolumn{2}{c}{\textbf{Counting Type Question Patterns}}\\
        \hline
        How many [E1] are [R] the [E2]? & How many tables are left for the 'Result and Discussion'?  \\ 
        What is the number of [E1] [R] the [E2]? & What is the number of tables below the 'Background \& Summary'?  \\ 
        How many [E1] can you find on the [R] of [E2]?  & How many figures are upper the 'Discussion'?  \\
        Count the number of [E1] on the [R] of [E2]. & Count the number of figures below 'Material and methods'.  \\
        When you check this page, how many [E1] can you find on the [R] of [E2]? & When you check this page, how many tables can you find on the top of 'Background'?  \\ \hline
        Can you find [num] [E](s) on the page? & Can you find 2 table(s) in the page?  \\ 
        Does this page include [num] [E](s) & Does this page include 2 figures?  \\ 
        Confirm if there are [num] [E](s) on this page. & Confirm if there are 1 table(s) in this page.  \\
        Are there [num] [E](s) on this page?  & Are there 3 figure(s) in this page?   \\ 
        Is there only [num] [E](s) on this page?  & Is there only 2 table(s) in this page?   \\ \hline
        How many [E]s on this page? & How many tables in this page?  \\ 
        When you check this page, how many [E]s are on this page? & When you check this page, how many tables are on this page?  \\ 
        What is the number of [E]s on this page? & What is the number of figures on this page?  \\
        How many [E]s can be found on this page? & How many figures can be found on this page? \\ \hline
    
    \end{tabular}
    \end{adjustbox}
        \end{center}
    \caption{Task A question pattern templates with corresponding example questions.}
    \label{tab:taska_templates}
\end{table*}

\begin{table*}[!ht]
\begin{center}
    \centering
    \begin{adjustbox}{max width=\textwidth}
    \begin{tabular}{p{0.8\textwidth}p{0.8\textwidth}}
    \hline
    \multicolumn{1}{c}{\textbf{Question Pattern}} &
    \multicolumn{1}{c}{\textbf{Question Example}} \\
    \hline
    \multicolumn{2}{c}{\textbf{Structural Understanding}}\\
    \hline
        What is the [turn] section in this page? & What is the last section in this page?  \\ 
        Can you describe the [turn] section of this page? & Can you describe the first section of this page?  \\ 
        What does the [turn] section include in this page? & What does the last section include in this page?  \\ 
        What is the main contents of the [turn] section in this page? & What is the main contents of the first section in this page?  \\ 
        When you check the [turn] section of this page, what information can you get? & When you check the last section of this page, what information can you get?  \\ \hline
        What is the [pos] section about? & What is the top section about?  \\ 
        What is the [pos] of the page about? & What is the left of the page about?  \\ 
        What is the topic of [pos] section? & What is the topic of bottom section?  \\ 
        Can you describe the main topic of the [pos] section? & Can you describe the main topic of the right section?  \\ 
        When you check the [pos] of this page, what information can you get? & When you check the bottom of this page, what information can you get? \\ \hline
    
    \hline
    \multicolumn{2}{c}{\textbf{Object Recognition}}\\
    \hline
    What is the [E] on the [pos] of the page? & What is the table on the top of the page?  \\ 
        What is the [pos] [E] about? & What is the bottom table about?  \\ 
        Can you describe the [E] on the [pos] of the page?  & Can you describe the figure on the bottom of the page?   \\ 
        What information does the [pos] [E] contain? & What information does the left figure contain?  \\ 
        When you check the [pos] [E], what information can you get? & When you check the top table, what information can you get? \\ \hline
        
    \end{tabular}
    \end{adjustbox}
        \end{center}
    \caption{Task B question pattern templates with corresponding example questions.}
    \label{tab:taskb_templates}
\end{table*}

\begin{table}[!ht]
    \begin{center}
    \centering
    \begin{adjustbox}{max width=\textwidth}
    \begin{tabular}{p{0.8\textwidth}p{0.8\textwidth}}
    \hline
    \multicolumn{1}{c}{\textbf{Question Pattern}} &
    \multicolumn{1}{c}{\textbf{Question Example}} \\
    \hline
    \multicolumn{2}{c}{\textbf{Child Relation Understanding}} \\
    \hline
       What does the [E] include? & What does the Introduction include?  \\ 
        What is the [E] about? & What is the Competing interests about?  \\ 
        What subsections are in the [E]? & What subsections are in the 2. Clinical Presentation?  \\
        What subsections can be found in the [E]? & What subsections can be found in the Materials and methods?  \\ 
        When you check the [E], which subsections are included?  & When you check the Methods, which subsections are included? \\ \hline
    
    \hline
    \multicolumn{2}{c}{\textbf{Parent Relation Understanding}} \\
    \hline
    Which section does describe the [E] ? & Which section does describe the Table 3?  \\ 
        Which section does include the description of the [E]? & Which section does include the description of the Table 2?  \\ 
        Name out the section that include the [E].  & Name out the section that include the Table 2.   \\ 
        Where can you find the [E]? & Where can you find the Table 2?  \\ 
        When you search for the description of [E], which sections do you need to check? & When you search for the description of Figure 1, which sections do you need to check?  \\ \hline
        Which section does include the [E]? & Which section does include the 'Corwin HL et al,2009'?  \\ 
        Which section does cite the [E]? & Which section does cite the 'Wang C et al,2017'?  \\ 
        Where is the [E] cited in the document? & Where is the 'Horner KC et al,2005' cited in the document?  \\ 
        Where can [E] be found in the document? & Where can 'Guan KL et al,1991' be found in the document?  \\ 
        When you search for the citation of [E], which sections can you find it?  & When you search for the citation of 'Zhang Z et al,2013', which sections can you find it? \\ \hline
        
    \end{tabular}
    \end{adjustbox}
    \end{center}
    \caption{Task C question pattern templates with corresponding example questions.}
    \label{tab:taskc_templates}
\end{table}

\section{Human Evaluation Details}
\begin{figure}[t]
 \centering
 \includegraphics[width=0.8\linewidth]{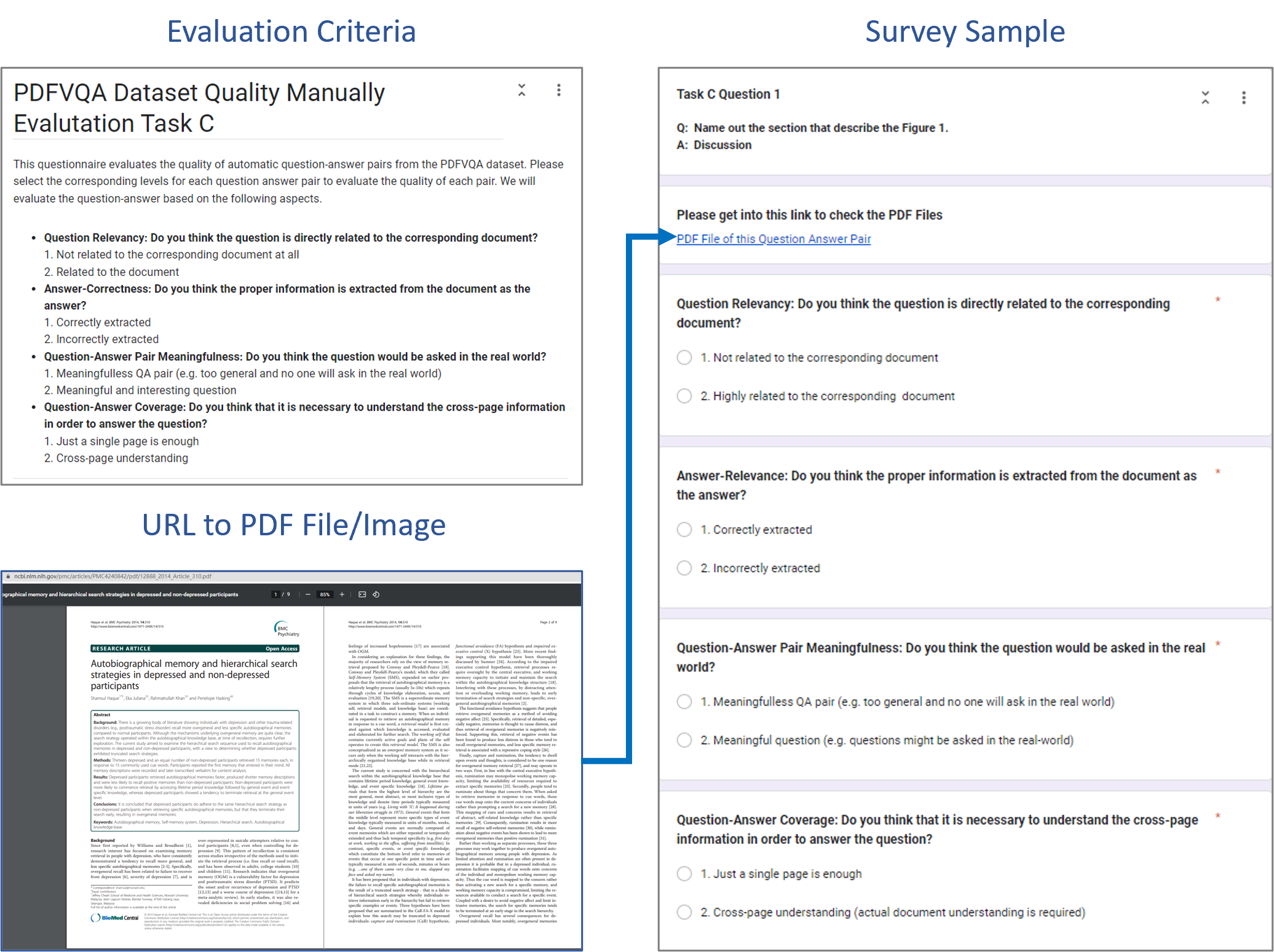}
 \caption{A human evaluation sample with evaluation criteria of Task C. Task A and B have a similar style as Task C.}
 \label{fig:survey_sample}
\end{figure}

We randomly selected 30, 30 and 40 question-answer pairs from Task A, Task B and Task C, respectively and put them with the related document page images or file links in the google forms (An example of Task C can refer to Figure~\ref{fig:survey_sample}). For each task, raters need to check each generated question-answer pair together with the attached document page or file to determine whether the question-answer pairs meet the requirements of three aspects, \textit{Relevance}, \textit{Correctness}, \textit{Meaningfulness}. For example,  for a given question in Figure~\ref{fig:survey_sample}, "Name out the section that describes Figure 1", raters need to first go through the entire document to check whether the document has Figure 1 and then check which sections provide the description of that figure to compare with the provided answer. Finally, raters are required to determine whether this question will be asked in the real world. We show the detailed definition of each aspect to ensure raters can understand the evaluation metrics of each criterion at the beginning of the questionnaire of each task, as Figure~\ref{fig:survey_sample} shows. 

\section{Additional Dataset Analysis}
\subsection{Distribution of Question Length}
\begin{figure}[t]
 \centering
 \includegraphics[width=\linewidth]{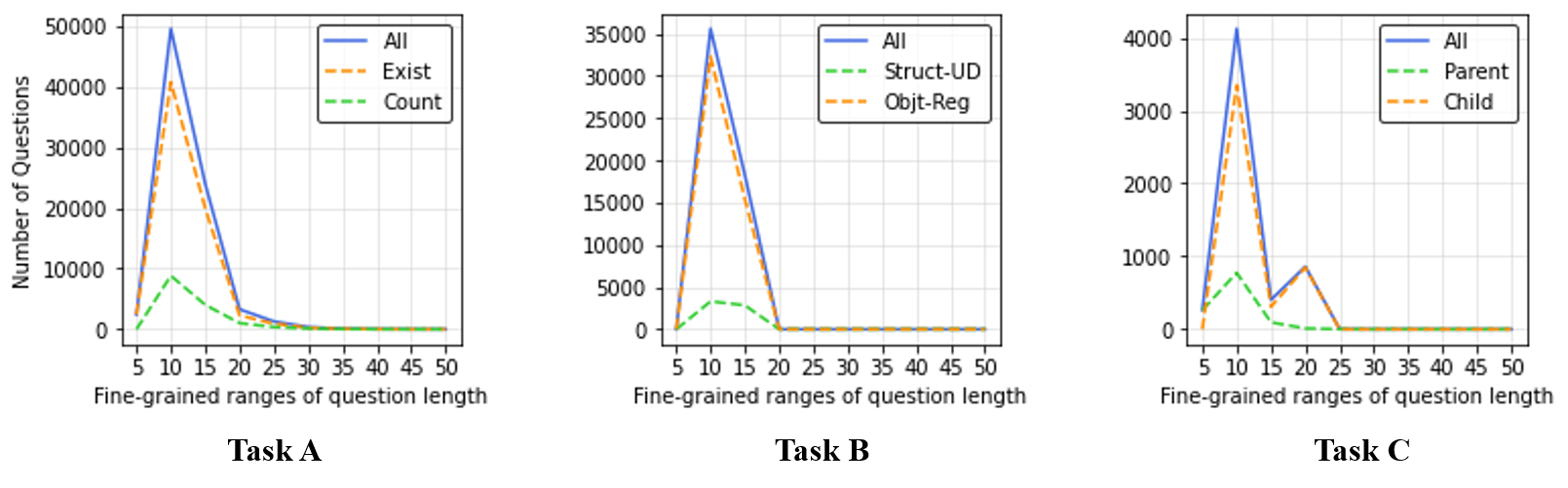}
 \caption{Question length distribution of Task A, B and C}
 \label{fig:question_len}
\end{figure}

We show the distribution of question length of each task in Figure~\ref{fig:question_len}. The average question length for Task A, B and C are 25, 10 and 15, respectively.

\section{Baseline Details}
\subsection{Baseline Descriptions}
\begin{itemize}
    \item \textbf{M4C}~\cite{hu2020iterative} applies the multimodal transformer, which takes into the question embedding, OCR token embedding and image object features as inputs, and iteratively decodes the answers over the combined answer space of OCR tokens and the fixed answer list. 
    \item \textbf{VisualBERT}~\cite{li2019visualbert} is a pretrained vision-and-language model that passes the sequence of text and object region embeddings to a transformer to get the integrated vision-and-language representations. 
    \item \textbf{LXMERT}~\cite{tan2019lxmert} applies three transformer encoders to encode the text embeddings, object region embeddings and the cross-modality learning between texts and image features. 
    \item \textbf{ViLT}~\cite{kim2021vilt} operates linear projection over image patches to get a sequence of image patch representations and input to the transformer encoder together with the text embeddings to get a pretrained vision-and-language model. 
    \item \textbf{BERT}~\cite{devlin2019bert} is a pretrained language model that applies the structure of a multi-layer bidirectional transformer encoder. We used only the textual features from document pages as the inputs.
    \item  \textbf{LayoutLMv2}~\cite{xu2021layoutlmv2} is a pre-trained model to operate on the position and textual features of document elements and generate the integrated representations that can be used for downstream document-related tasks. 
\end{itemize}

\subsection{Baseline Setup}
\begin{itemize}
    \item \textbf{M4C} applied multiple transformer layers, learning question embeddings, image object features and OCR token features in the common embedding space and iteratively decoding answer tokens from a fixed answer space or the OCR tokens in the image. The OCR tokens are encoded in rich representations, including the textual embedding of each token, appearance features of the token region on the image, Pyramidal Histogram of Characters (PHOC) features and the location features. We evaluated all three tasks with M4C but slightly modified the inputs and output layer to suit document-based VQA. Firstly, since the number of OCR tokens is much larger in PDF documents than that in real-life scenes, instead of inputting the features of all OCR tokens in the page, we used the BERT [CLS] token features to represent the sequence of textural contents in each document element region and took them together with the question embedding and the visual features of each document element region as the input sequence to the multi-layer transformer. Secondly, in the decoding part, Task B and C, we used the $d$-dimensional representations for the index numbers of the corresponding document element region in the page and generated the scores through the dynamic pointer network to predict the index number of document element region over the list of document element region index numbers. For applying M4C to Task A, we set fixed answer space as the decode inputs and put the pointer network on top to get a final prediction.
    \item \textbf{BERT}, \textbf{LayoutLM2} are used only for Task A and B because the inputs of both models are question and context token level information with the 512 maximum limitations. For multi-page documents, the number of tokens is normally much higher than 512 tokens, which means those two models can only catch the first-page context information. In this case, we did not select those two models for conducting Task C tests. For both Task A and B, we directly extract 768-dimension [CLS] token embedding and feed it into classifiers for predicting the corresponding answer or object sequential index.  
    \item \textbf{VisualBERT}, \textbf{LXMERT} can process visual features of document layout elements extracted from pretrained ResNet101-Res5. After we feed those raw object-level visual features and question tokens into those vision-language pretrained models, we extract the enhanced visual representation of document layout elements and feed them into a pointer network to get final scores for predicting corresponding answers for all three tasks.
    \item \textbf{ViLT} is directly applied for conducting Task A and B by using the provided feature extractor and pre-trained 
    \item \textbf{ViltForQuestionAnswering} model to predict the corresponding answer based on input questions and image patch features. For addressing task C, we concatenate all document pages into an image pixel matrix and feed into the feature extractor to extract image patch features for feeding forward pass. The outputs pass through a Sigmoid layer instead of the Softmax function adopted by other tasks for backward propagation in the training stage and answer prediction in the inference stage.
\end{itemize}

\section{Implementation Detail}
Dimension for the visual features of each document element region $d_f$ is 2048. The activation function used in GCN is Tanh. The GCN is trained with AdamW optimizer and 0.0001 learning rate for 10 epochs. Each question token is encoded into a 768-dimension fine-tuned on the BERT-base model. Our model utilized a $6$ layers transformer encoder and a $4$ layers transformer decoder with $12$ heads and 768-dimension hidden size. The maximum numbers for input question tokens and objects (document layout elements) are 50 and 25, respectively, for Task A and B and 50 and 400 for Task C. For a fair comparison, epoch times are selected as 5, 10, and 20 for all Task A, B and C models, respectively. All the experiments are conducted on 51 GB Tesla V100-SXM2 with CUDA 11.2.
\end{document}